\documentclass[11pt]{article}
\usepackage{coling2020}
\usepackage{times}
\usepackage{url}
\usepackage{multirow}
\usepackage{graphicx}
\usepackage{latexsym}
\usepackage{algorithm}
\usepackage{algpseudocode}
\usepackage{adjustbox}
\usepackage[dvipsnames]{xcolor}
\usepackage{soul}
\colingfinalcopy

\title{Biased TextRank: Unsupervised Graph-Based Content Extraction}

\author{Ashkan Kazemi, Ver{\'o}nica P{\'e}rez-Rosas, Rada Mihalcea \\
  Department of Computer Science \& Engineering \\
  University of Michigan, Ann Arbor \\
  \texttt{\{ashkank, vrncapr, mihalcea\}@umich.edu}}

\date{}

\begin{document}

\maketitle

\begin{abstract}
We introduce Biased TextRank, a graph-based content extraction method inspired by the popular TextRank algorithm that ranks text spans according to their importance for language processing tasks and according to their relevance to an input ``focus.'' Biased TextRank enables focused content extraction for text by modifying the random restarts in the execution of TextRank. The random restart probabilities are assigned based on the relevance of the graph nodes to the focus of the task. 
We present two applications of Biased TextRank: focused summarization and explanation extraction, and show that our algorithm leads to improved performance on two different datasets by significant ROUGE-N score margins. Much like its predecessor, Biased TextRank is unsupervised, easy to implement and orders of magnitude faster and lighter than current state-of-the-art Natural Language Processing methods for similar tasks.
\end{abstract}

\section{Introduction}
Content and information extraction 
are central to many Natural Language Processing (NLP) tasks, from question answering~\cite{rajpurkar-etal-2018-know,reddy-etal-2019-coqa} to text summarization~\cite{hermann2015teaching,dang2005overview} and beyond. While the state-of-the-art solutions for these tasks mainly rely on training neural network architectures on very large datasets, there have been questions around the sustainability of these solutions and their effects on the environment. As highlighted in work by~\newcite{strubell-etal-2019-energy}, training one large transformer-based model produces approximately four times more $CO_2$ emissions than a car in its lifetime. These considerable negative environmental outcomes call for lighter and less resource-intensive alternative methods.

TextRank~\cite{mihalcea-tarau-2004-textrank} is a light-weight unsupervised graph-based content extraction algorithm that was initially designed for summarization and keyword extraction applications. Since its introduction, it has been adapted and used in numerous other applications and settings, including opinion mining \cite{petasis2016identifying,deguchi-yamaguchi-2019-argument}, credibility assessment \cite{balcerzak2014application} and lyrics summarization \cite{son2018music}, among others. Most recently, TextRank has been included in the latest release of the popular spaCy library.\footnote{\url{https://spacy.io/universe/project/spacy-pytextrank}} There have been online tutorials 
and updating studies~\cite{barrios2015variations} that demonstrate TextRank's relevance years after its initial release. 

Some of the TextRank extensions that have been proposed in recent years rely on the idea of personalized (or topic-sensitive) PageRank~\cite{haveliwala2003topic} and its successor algorithms. For instance, PositionRank~\cite{florescu2017position} changed the TextRank rankings to account for the position of candidate words in the input document, and showed that this position-aware algorithm led to improvements in keyword extraction over TextRank and over several other baselines.

In this paper, we introduce Biased TextRank, which relies on  document representation models and similarity measures that enable capturing meaning closeness between graph nodes and a target (focus) text. While we demonstrate the usefulness of our approach on two applications -- focused summarization and explanation extraction -- we believe  this approach is generalizable to other applications that require content extraction and/or content ranking.

The paper makes the following three main contributions:
\begin{enumerate}
    \item \textbf{Biased TextRank}: We introduce an unsupervised graph-based algorithm for focused content-extraction that does not require training data, is fast, resource-efficient and easy to implement and fine-tune. Biased TextRank is language agnostic, in the sense that as long as document embedding models exist for a language, Biased TextRank can be directly applied. With the recent emergence of technologies like LASER~\cite{artetxe2019massively} with pretrained language embeddings for 100+ languages, such representations are readily available for many languages.
    \item \textbf{Evaluation and extensive analyses of Biased TextRank}: We show the effectiveness of Biased TextRank through experiments on two tasks: focused summarization and explanation extraction. We also perform an ablation study to show the effects of the TextRank damping factor and the similarity threshold parameters, providing insight on how Biased TextRank parameters should be tuned.
    \item \textbf{Focused summarization dataset}: We introduce and make available a novel dataset for focused summarization consisting of transcripts of the U.S. presidential debates during the past 40 years, alongside articles from both Democrat and Republican media summarizing the events of the debates.
\end{enumerate}

The remainder of the paper is structured as follows: section~\ref{related_work} covers prior work. In section~\ref{alg_desc} we provide a step-by-step description of our proposed algorithm. Throughout section~\ref{experiments}, we describe two applications of Biased TextRank: focused summarization (section~\ref{focused_summarization}) and explanation extraction (section~\ref{explanation_extraction}). We dedicate section~\ref{sec:ablation} to an ablation study to understand the role played by the similarity threshold and damping factor parameters on Biased TextRank. Finally, we discuss our findings and possible future directions in section~\ref{discussion} and conclude the paper in section~\ref{conclusion}.

\section{Related Work}\label{related_work}
\subsection{TextRank}
Inspired by PageRank~\cite{ilprints422}, the TextRank algorithm~\cite{mihalcea-tarau-2004-textrank} is a 
content extraction algorithm that represents texts as graphs for sentence and keyword extraction purposes and uses the PageRank algorithm to rank sentences or keywords. Since TextRank was first released, it has been applied to tasks such as summarization~\cite{mallick2019graph,barrios2015variations,son2018music}, keyword extraction~\cite{wen2016research,jianfei2016using},  opinion mining~\cite{petasis2016identifying,deguchi-yamaguchi-2019-argument}, credibility assessment \cite{balcerzak2014application} and others. 

Among these, the work closest to ours are presented in~\cite{wan2008using} and~\cite{florescu2017position}. In~\newcite{wan2008using} the author explored inter and intra-document relationships in generic and topic-focused multidocument summarization. They used TextRank and a combination of inter and intra-document edge weighting mechanisms alongside a diversity penalty to solve DUC 2002-2005 multidocument summarization tasks. Their encoding of ``focus'' into the algorithm is similar to our approach, except that they implemented it using tf-idf vectors and targeted the specific task of multidocument summarization. In~\newcite{florescu2017position} the authors proposed ``PositionRank,'' a keyword-extraction method based on TextRank and personalized PageRank. In their work, they bias the TextRank scores based on how early the keywords appear in the input document and their method is designed for keyword extraction only. Although our method also relies on biasing the TextRank scores, there are two important differences. First, Biased TextRank provides a different solution to the underlying content extraction problem as it uses contextual embeddings and similarities that allow for a topical focus. Second, it is not limited to keyword extraction or multidocument summarization and can be used for a wide variety of applications.

\subsection{Focused Summarization}
Although focused summarization has not been widely studied within the NLP community, query-focused or query-biased summarization is a known problem in the context of Information Retrieval~\cite{wang2007learning,metzler2008machine,zhao2009using}. \newcite{wang2007learning} proposed two extractive query-biased summarization methods (classification and ranking-based) for web page summarization. They extracted features from both the content and context of a web page and feed them to an SVM that solves both the classification and ranking problem formulations. More recently in \newcite{cao-etal-2016-attsum} the authors proposed AttSum, a system that leverages joint learning of query relevance and sentence salience ranking, the two main modules of query-focused summarization and achieve competitive results on the DUC~\cite{dang2005overview} datasets.
While related work has been published in NLP venues~\cite{daume-iii-marcu-2006-bayesian}, query-focused summarization has been mainly studied by the Information Retrieval community.

\subsection{Explanation Extraction}
Model explainability~\cite{poursabzi2018manipulating,lundberg2017unified} and natural language explanation extraction and generation~\cite{kumar-talukdar-2020-nile,thorne-etal-2019-generating} are broad and important topics of ongoing research within the AI and NLP communities. However explanation extraction in the context of fact-checking and misinformation detection has remained relatively understudied. In \newcite{atanasova-etal-2020-generating}, the authors address the task of extracting fact-checking explanations, in which statements documenting the veracity of a fact-checked statement are used to derive a short summary explanation. The authors propose a BERT~\cite{devlin-etal-2019-bert} based sentence selection model that 
identifies top relevant sentences from the input as candidate explanations. In similar context, highlighting natural language explanations for fact-checking and misinformation detection applications has been studied within the research community~\cite{lu-li-2020-gcan,popat-etal-2018-declare}.

\section{Biased TextRank}\label{alg_desc}
Biased TextRank builds upon the original TextRank algorithm, but changes how random restart probabilities are assigned, therefore giving higher likelihood to the nodes that are more relevant to a certain  ``focus'' of the task. 

\subsection{Node Scoring with Random Restart Probabilities}
\paragraph{TextRank Node Scoring.} TextRank operates on graphs that are built from natural language texts. For instance, in the original TextRank application, the graphs are built from sentences in a text, or from individual words. The text spans are connected through links that are extracted from text, which reflect the strength of the relation between those spans. For instance, sentences can be linked by their similarity, or words can be linked by their proximity in the text. Assuming a graph representation with nodes $V_i$ and the edges between nodes having a weight $w_{ij}$, TextRank uses the following formula to iteratively update the $TextRank$ score of a node:\\

\noindent
\begin{center}
\begin{math}
TextRank(V_{i}) =  (1-d) + d* \sum\limits_{V_j \in In(V_{i})}{\frac{w_{ji}}{\sum\limits_{V_k \in Out(V_{j})}{w_{jk}}}TextRank(V_{j})}
\end{math}
\end{center}

\noindent where d is a damping factor typically set to 0.85.

\paragraph{Biased TextRank Node Scoring.} In TextRank, each node has an equal random restart probability, and therefore all the nodes are treated equally during the application of the algorithm. 
Biased TextRank however operates on assigning these random restart probabilities to favor a specific  focus. When executing the algorithm, the nodes that have a high random restart likelihood will have a higher chance of being reached during the random jump. Therefore, the ranking algorithm is changed to:\\

\noindent
\begin{center}
\begin{math}
BiasedTextRank(V_{i}) =  BiasWeight_i * (1-d) + d* \sum\limits_{V_j \in In(V_{i})}{\frac{w_{ji}}{\sum\limits_{V_k \in Out(V_{j})}{w_{jk}}}BiasedTextRank(V_{j})}
\end{math}
\end{center}

\noindent where $BiasWeight_i$ is set to a value that reflects the relevance of the node $V_i$ for the focus of the task, and the damping factor d is set as before to 0.85. We further explore the role of the damping factor in the effectiveness of the Biased TextRank algorithm in Section \ref{sec:ablation}.

\subsection{Biased TextRank Algorithm} 

The algorithm starts with a document, and produces a ranking over text spans according to the biased TextRank formula shown earlier. The input document is first
parsed into chunks that are then embedded into vectors to facilitate computation. 
These vectors constitute the 
nodes of the graph, which are then used to determine a ranking for the sentences. The focus (or bias) of the task is also embedded, and used to calculate the bias weights. After ranking, the top $K$ ranked sentences are selected and returned as a result. 

Algorithm~\ref{pseudocode} illustrates this procedure. We use matrix representations such that all vertices are processed in one step. We discuss each step in detail in the following subsections.

\begin{algorithm}[ht]
  \begin{algorithmic}[1]
    \Procedure{BiasedTextRank}{$Document, Bias, SimilarityThreshold, DampingFactor$} 
        \State Sentences = PARSE(document)
        \State SentenceEmbeddings = EMBED(Sentences)
        \State BiasEmbedding = EMBED(Bias)
        \State SimilarityMatrix = GRAPH CONSTRUCTION(SentenceEmbeddings,SentenceEmbeddings)
        \State BiasWeights = RANDOM RESTART PROBABILITIES(BiasEmbedding,SentenceEmbeddings)
        \For{each Row in SimilarityMatrix}
            \For{each Cell in Row}
                \If{Cell.value $<$ SimilarityThreshold}
                    \State Cell.value = 0
                \EndIf
            \EndFor
        \EndFor
        \State SimilarityMatrix = \\
        \hskip3em $DampingFactor * SimilarityMatrix + (1 - DampingFactor) * BiasWeights$
    \State RETURN RandomWalkWithRestart(SimilarityMatrix)
    \EndProcedure
    \\
    \Procedure{RandomWalkWithRestart}{$Matrix$}
        \State n = Matrix.len
        \State Ranks = new Array(n).INITIALIZE($1/n$)
        \For{as long as Ranks has not converged}
            \State Ranks = $Matrix^T \cdot Ranks$
        \EndFor
        \State RETURN Ranks
    \EndProcedure
\end{algorithmic}
\caption{Pseudocode for Biased TextRank.}
\label{pseudocode}
\end{algorithm}

\paragraph{PARSE.}
Biased TextRank extracts relevant context from input documents by ranking document pieces. In order to do this, we need to parse the documents into those pieces. For instance, if the algorithm is used for sentence extraction, we parse the input into sentences. If it is to be used for keyword extraction, we parse the input into tokens. 

\paragraph{EMBED.}
Transforming documents into graphs requires mathematical representations of the nodes of the graph. This mathematical representation will enable similarity comparison between nodes, an integral part of the TextRank algorithm. With recent advances in contextual embedding technologies, we find Sentence-BERT (SBERT)~\cite{reimers-gurevych-2019-sentence} to be a good model to embed English texts. For non-English sentence embedding, contextual embedding models like LASER~\cite{artetxe2019massively} are useful. Word embedding models like Word2Vec~\cite{mikolov2013distributed} can similarly be used in the case of keyword extraction. After embedding document pieces $DP_i$, $i = 1..n$ into embedding vectors $E_i$, $i = 1..n$ of fixed length, we can build a representative graph of the input document.

\paragraph{GRAPH CONSTRUCTION.}
To build a graph representation of the input, we follow the same graph building strategy as in the original TextRank algorithm. For sentence extraction, the process is as follows: Each sentence embedding $SE_i$ is represented as a node $V_i$ in a graph $G_D$ of the input document. We add an edge 
$E_{ij}$ connecting nodes $V_i$ and $V_j$, if $SimilarityMeasure(SE_i, SE_j) > SimilarityThreshold$. The weight $w_{ij}$ of $E_{ij}$ equals $SimilarityMeasure(SE_i, SE_j)$. We use cosine similarity as our $SimilarityMeasure$. We discuss the selection of the $SimilarityThreshold$ in section~\ref{sec:ablation}.

\paragraph{RANDOM RESTART PROBABILITIES.}
Assigning random restart probabilities to nodes is key for making Biased TextRank work. Similar to the topic-sensitive PageRank algorithm~\cite{haveliwala2003topic}, this is achieved by assigning higher restart probabilities to nodes that are most similar to the focus of the task. We use a short text describing the focus of the content extraction to determine the similarity between the nodes and the task. We transform the description into a fixed-length embedding vector using the EMBED procedure (bias embedding vector) and calculate its similarity to the nodes. The higher the similarity (obtained using cosine similarity or any other similarity measure) between a node and the bias embedding vector, the higher restart probability is assigned to that node.

\section{Experiments}\label{experiments}
We conduct two main experiments to explore the ability of Biased TextRank to perform focused content extraction.

\subsection{Experimental Settings}
We implement Biased TextRank using the NLTK library and SBERT in Python. For sentence embedding retrieval, we use the pretrained, base SBERT model. We run our experiments on a machine using one Nvidia 1080 Ti GPU and the GPU is only used to make embedding retrieval faster. A run of Biased TextRank for large documents on a graph with approximately 1,000 nodes takes an average 1.6 seconds to complete. This measurement also includes the embedding retrieval time.\footnote{We used a GPU to facilitate embedding retrieval. For the same large graph benchmark, Biased TextRank takes 51.5 seconds on average to complete on a laptop with 4 CPU cores and 8GBs of RAM, 50.8 of those going to embedding retrieval.}

Since all of our experiments focus on sentence extraction, we use the sentence tokenizer from the NLTK~\cite{bird-loper-2004-nltk} library. 
During our evaluations we use the ROUGE~\cite{lin-2004-rouge} as the main performance metric.

\subsection{Focused Summarization}\label{focused_summarization}
Focused summarization, much like query-focused summarization (its counterpart in information retrieval), aims to generate summaries for an input text with a given focus. 


To evaluate the applicability of Biased TextRank for extracting focused summaries, we collected a dataset of  news reportage from Democrat and Republican media's interpretations of the U.S. presidential debates from 1980 to 2016. We use the collected news reportage that summarize the events of the debates and apply Biased TextRank to reproduce the biased interpretations of Democrat and Republican media. The New York Times online public archives are the source of our Democrat summary references. For Republican debate coverage, we collect reportage from Fox News, The New York Post and Houston Chronicle. We also collected debate transcripts from debates.org, a public resource by the U.S. Commission on Presidential Debates. Since it is difficult to find news covering older debates, we could not find a number of articles that cover presidential debates of the 1970s and 1960s from either side. General statistics for the collected dataset of U.S. presidential debate news coverage are presented in table~\ref{summarization_dataset_stats}.

\begin{table}[t]
\begin{center}
\begin{tabular}{ l  c  c  c } 
\hline
 & \#documents & avg \#tokens & std \#tokens \\
\hline
Democrat & 26 & 2130 & 406 \\ 
Republican & 22 & 1087 & 281 \\ 
Transcripts & 33 & 18868 & 4708 \\ 
\hline
\end{tabular}
\caption{Focused Summarization Dataset Statistics}
\label{summarization_dataset_stats}
\end{center}
\end{table}

 To generate the focused summaries we use debates' transcripts and a fixed bias description for each side. We pick the Republican bias text from the opening paragraphs in the Republican party Wikipedia page that describes party values. For the Democrat bias text, we choose the headlines of their most recent party platform document.\footnote{\url{https://democrats.org/wp-content/uploads/sites/2/2019/07/2016_DNC_Platform.pdf}} After running Biased TextRank on the parsed debate transcripts, we pick the top 20 ranked sentences as the focused summary. We also obtain unfocused summaries of the debates using TextRank. Our implementation of TextRank is identical to Biased TextRank, with the difference that each node gets an equal random restart probability.

Table~\ref{summarization_results} presents the results when comparing generated summaries against the corresponding Democrat and Republican ground truths. 
As observed, the focused summaries outperform unfocused summaries on both sides in capturing a biased overview of the debates. For the Democrat summary references, Biased TextRank has a gain of 13.05, 2.3 and 4.52 ROUGE-1, ROUGE-2 and ROUGE-L F1 scores respectively over TextRank. Similar differences of 11.8 ROUGE-1, 2.47 ROUGE-2 and 3.72 ROUGE-L F1 scores emerge in the Republican ground truth as well. We attribute the performance gap to the attention of Biased TextRank to the underlying biases already existing in the ground truth text. 

Overall, the experiments show that focused summaries produced by Biased TextRank meaningfully improve over normal summaries when compared against a biased reference. We believe Biased TextRank is a better fit than conventional extractive summarization methods when there is a clear focus or bias required in the desired summary.

\begin{table}[t]
\begin{center}
\begin{tabular}{ c  c  c  c  c } 
\hline
Party & Method & ROUGE-1 & ROUGE-2 & ROUGE-L \\
\hline
\multirow{2}{5em}{Democrat} & TextRank & 17.04 & 3.54 & 16.83 \\ 
 & BiasedTextRank & \textbf{30.09} & \textbf{5.84} & \textbf{21.35} \\ 
\hline
\multirow{2}{5em}{Republican} & TextRank & 21.86 & 3.38 & 18.39 \\ 
 & BiasedTextRank & \textbf{33.66} & \textbf{5.85} & \textbf{22.11} \\ 
\hline
\end{tabular}
\caption{Results for Biased TextRank application in focused summarization are evaluated using the ROUGE-N F1 score. The party column determines the set of ground truths that summaries are compared against; Democrat refers to the New York Times articles covering the debates and Republican refers to a collection of ground truth text collected from Fox News, the New York Post and Houston Chronicle.}
\label{summarization_results}
\end{center}
\end{table}

\subsection{Explanation Extraction}\label{explanation_extraction}
Introduced as ``explanation generation'' for fact-checking by~\newcite{atanasova-etal-2020-generating}, this task focuses on extracting explanations from articles elaborating on the veracity of statements in the PolitiFact-based LIAR-PLUS dataset~\cite{alhindi-etal-2018-evidence}. The dataset consists of 2,533 data points split into 1,278 validation and 1,255 test. Each data point consists of a statement, its veracity (e.g., true, false, mostly-true), a detailed article justifying the assigned veracity of the statement by fact-checkers, and a closing paragraph summarizing the explanation of the verdict. The goal is to extract the closing statement (explanation) from the lengthy justifying article. Table~\ref{table:example} shows an example of explanation extraction on this dataset when using Biased TextRank.

We designate the justification article as the input text and use the statement to be fact-checked as the bias text to be fed into Biased TextRank. Similar to the \newcite{atanasova-etal-2020-generating} system, we pick the top 4 ranked sentences as the extracted explanation.
We compare the explanation extraction performance of Biased TextRank with two unsupervised baselines: the Lead-4 baseline from~\newcite{atanasova-etal-2020-generating}, which takes the leading 4 sentences of the input as the explanation; and TextRank, which computes an extractive summary of the fact-check report for an explanation. 
While~\newcite{atanasova-etal-2020-generating} introduced a supervised method trained on 10,146 instances, achieving 35.70 ROUGE-1, 13.51 ROUGE-2 and 31.58 ROUGE-L F1 scores on the LIAR-PLUS test set, we believe the results of their system are not directly comparable to ours, given our fully unsupervised setting. 

 The results for these experiments are presented in table~\ref{explanation_extraction_results}. 
As observed, Biased TextRank outperforms both unsupervised baselines by at least 2.92 ROUGE-1, 2.97 ROUGE-2 and 1.94 ROUGE-L F1 scores on the validation set and 2.79 ROUGE-1, 2.97 ROUGE-2 and 1.84 ROUGE-L F1 scores on the test set. We believe these improvements demonstrate Biased TextRank's effectiveness in extracting explanatory supporting sentences for a given claim as an unsupervised and lightweight method.

\begin{table}[t]
\small
\begin{center}
\begin{tabular}{| p{0.95\columnwidth} |} 
\hline
\textbf{Claim:} ``Nearly half of Oregon's children are poor.'' \\
\hline
\textbf{Fact-Check Report:} ``With the State Board of Higher Education handing oversight of Oregon's universities to independent boards, Jim Francesconi, one of the state board members, recently took to The Oregonian's opinion pages to note a few of the issues the new custodians will have to deal with. Among them he said, and most importantly, education has to be accessible. ``Oregon,'' he wrote, ``must demonstrate that working people and poor folks can still make it in America. Education after high school is the way, but it is out of reach for many children, especially in rural Oregon. \hl{Nearly half of Oregon's children are poor.}'' It was the line about the percentage of poor children in the state that caught one Oregonian reader's attention. Oregon is hardly a rich state -- particularly when the national economy itself is down and out -- but nearly half? That seemed a stretch. We agreed with our reader -- it was worth looking into. Our first call was to Francesconi to see where he got his figures. He said the information came from a 2012 report...According to that report, ``\hl{nearly 50\% of children are either poor or low-income.}'' Francesconi almost immediately realized his mistake. ``In retrospect, I wish I would have said poor or low income.''...\hl{there is a distinction between poor and low income as far as the U.S. government is concerned}. If you check the...Census information, you'll find that...\hl{23 percent of children in Oregon live in...below...poverty level while another 21 percent live in low-income families}. \hl{As far as the U.S. government is concerned, about a quarter of the state's children are poor, not half}... (redacted) \\
\hline
\textbf{Ground Truth:} So where does this leave us? Francesconi said in an opinion piece that ``nearly half of Oregon's children are poor.'' In fact, if you use federal definitions for poverty, about a quarter are poor and another quarter are low-income. But experts tell us that families that are described as low-income still struggle to meet their basic needs and, for all intents and purposes, qualify as poor. Be that as it may, Francesconi was referencing a report that used the federal definitions. \\
\hline
\textbf{Biased TextRank:} ``Nearly half of Oregon's children are poor.'' According to that report, ``nearly 50\% of children are either poor or low-income.'' Low income refers to families between 100 and 200 percent of the federal poverty level. As far as the U.S. government is concerned, about a quarter of the state's children are poor, not half. \\
\hline
\end{tabular}
\caption{An example of explanation extraction from the LIAR-PLUS dataset with important information highlighted. The first two rows show a claim and a redacted fact-check report of that claim, followed by actual and Biased TextRank extracted supporting explanations.}
\label{table:example}
\end{center}
\end{table} 

\begin{table}[t]
\begin{center}
\begin{tabular}{ l  c  c  c  c  c  c } 
\hline
\multirow{2}{4em}{Model} & \multicolumn{3}{c}{Validation} & \multicolumn{3}{c}{Test} \\
\cline{2-7}
 & ROUGE-1 & ROUGE-2 & ROUGE-L & ROUGE-1 & ROUGE-2 & ROUGE-L \\
\hline
Lead-4 & 27.92 & 6.94 & 24.26 & 28.11 & 6.96 & 24.38 \\
TextRank & 27.72 & 7.41 & 23.19 & 27.74 & 7.42 & 23.24 \\
\hline
Biased TextRank & \textbf{30.84} & \textbf{10.38} & \textbf{26.20} & \textbf{30.90} & \textbf{10.39} & \textbf{26.22} \\
\hline
\end{tabular}
\caption{Explanation extraction evaluations. The performance of our Biased TextRank unsupervised system is compared against two unsupervised baselines. }
\label{explanation_extraction_results}
\end{center}
\end{table}

\section{Ablation Study}\label{sec:ablation}
\begin{figure}[t]
\centering
\includegraphics[width=\columnwidth]{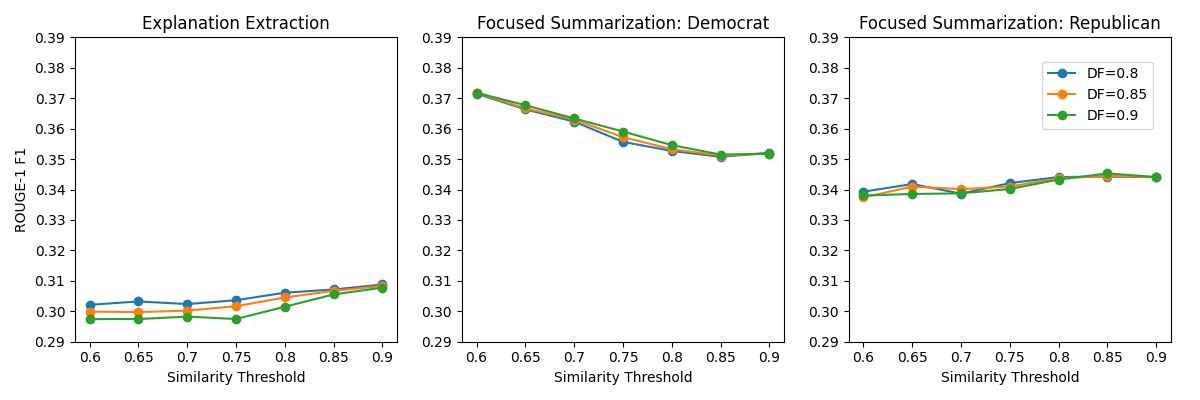}
\caption{Charts demonstrating ablation study results. Columns refer to experiments; the second and third columns are the two parts of the focused summarization experiment. DF refers to damping factor.}
\label{fig:ablation}
\end{figure}

To understand how the algorithm parameters affect Biased TextRank, we carry out an ablation study where we examine how the damping factor and the similarity threshold affect the rankings produced by Biased TextRank across tasks. The results of the study are presented in figure~\ref{fig:ablation}. The similarity measure (cosine similarity) and the document embedding model (SBERT) in this study are fixed. Also, while conducting this experiment, we increase the number of selected summary sentences from 20 to 30 to add more variance to our visualizations. 

We derive the following observations from the ablation study:
    (1) The damping factor, within suggested ranges found in the literature (0.8 to 0.9), has very limited effect on Biased TextRank for focused summarization and explanation extraction.
    (2) With the exception of the Democrat focused summaries, the variance of the similarity threshold does not significantly change the outcome of Biased TextRank. For the Democrat focused summarization experiment, a lower similarity threshold, which translates to a more dense graph representation of the document, yields better results.
    (3) 
    Given these results, we recommend setting the damping factor to 0.85 (or anywhere between 0.8 to 0.9) and the similarity threshold around 0.65 to obtain reasonable results.


\section{Discussion \& Future Work}\label{discussion}
In this paper, we showed that Biased TextRank is a promising method for focused content extraction. We believe that a written description of the focus of a content-extraction task is an intuitive way of operationalizing a corresponding solution. In our experience using Biased TextRank, we found that choosing the right bias text to direct the focus of content extraction is key in making Biased TextRank work. Like the explanation extraction task in~\ref{explanation_extraction}, sometimes the bias term comes as an input to the algorithm. However that is not always the case, as the focused summarization experiment (among other tasks) require manual selection of the bias term. In those environments it is important to choose biases that best reflect the intention of the task focus and be aware of short-comings of word embeddings. In our experience in the focused summarization experiment, we initially found our Republican and Democrat summaries to be more similar than desired. As we probed the summaries, we found that although different in essence, the two chosen bias terms were producing similar embedding vectors and we suspected the vagueness and word overlap in both bias terms were among the causes. After selecting more distinct and clear bias terms for each summary flavor, we observed more distinctions and desired properties in produced summaries. We are interested in studying and quantifying the effect of the bias text on the algorithm and develop a deeper understanding of how one should pick the right bias text for their goals.

We evaluate the use of Biased TextRank in focused sentence extraction for English texts only. However, we believe that Biased TextRank is language-agnostic in the sense that if we have the proper tools to parse and embed non-English documents, the algorithm will be directly applicable. With recent advances in multilingual contextual embedding technologies like LASER (which provides embeddings for more than 100 languages), we think it is possible to immediately apply it to languages other than English.

In future work we would like to explore the application of Biased TextRank beyond sentence extraction. For instance the ``term-set expansion'' task most recently tackled by~\newcite{kushilevitz-etal-2020-two}, in which an initial seed set of keywords are expanded by similar keywords found in a corpus, could be another application of our algorithm. The task can be modeled as a keyword extraction task similar to the example found in the original TextRank paper and the restart probabilities can be assigned based on node proximity to the initial seed set.
\section{Conclusion}\label{conclusion}
In this paper, we introduced Biased TextRank, an unsupervised graph-based algorithm for directed extraction of content from text. Biased TextRank is unsupervised, fast and resource-efficient, language-agnostic and easy to implement. We demonstrated its effectiveness on two tasks:  focused summarization and explanation extraction. 
For the first task, we collected a dataset of biased interpretations of the U.S. presidential debates by Democrat and Republican media 
and showed, through comparative experiments, that a Democrat or Republican-focused summary of the debates better captures those interpretations than a generic summary. For the second task, our explanation extraction experiments showed that Biased TextRank improved over the performance of two unsupervised baselines. 

In addition, we analyzed the effects of the damping factor and similarity threshold parameters on Biased TextRank through an ablation study and suggested parameter tuning guidelines for the algorithm.

Although we only demonstrated Biased TextRank effectiveness in two content extraction tasks, we believe that it can have a variety of natural language processing applications, similar to how TextRank has been used to address numerous tasks. With the satisfactory results of Biased TextRank, we find the approach to be a promising direction towards more sustainable content extraction solutions.

The Biased TextRank code as well as the focused summarization dataset we compiled are publicly available at \url{https://lit.eecs.umich.edu/downloads.html}.

\section*{Acknowledgments}

This material is based in part on work supported by the NSF (grant \#1815291), the John Templeton Foundation (grant \#61156), and by the University of Michigan through the MCubed program and the Educational Innovation Accelerator program. Any opinions, findings, conclusions, or recommendations in this material are those of the authors and do not necessarily reflect the views of the NSF,  the John Templeton Foundation, or the University of Michigan.

\bibliographystyle{coling}
\bibliography{anthology, extra_references}

\begin{thebibliography}{}

\bibitem[\protect\citename{Alhindi \bgroup et al.\egroup
  }2018]{alhindi-etal-2018-evidence}
Tariq Alhindi, Savvas Petridis, and Smaranda Muresan.
\newblock 2018.
\newblock Where is your evidence: Improving fact-checking by justification
  modeling.
\newblock In {\em Proceedings of the First Workshop on Fact Extraction and
  {VER}ification ({FEVER})}, pages 85--90, Brussels, Belgium, November.
  Association for Computational Linguistics.

\bibitem[\protect\citename{Artetxe and Schwenk}2019]{artetxe2019massively}
Mikel Artetxe and Holger Schwenk.
\newblock 2019.
\newblock Massively multilingual sentence embeddings for zero-shot
  cross-lingual transfer and beyond.
\newblock {\em Transactions of the Association for Computational Linguistics},
  7:597--610.

\bibitem[\protect\citename{Atanasova \bgroup et al.\egroup
  }2020]{atanasova-etal-2020-generating}
Pepa Atanasova, Jakob~Grue Simonsen, Christina Lioma, and Isabelle Augenstein.
\newblock 2020.
\newblock Generating fact checking explanations.
\newblock In {\em Proceedings of the 58th Annual Meeting of the Association for
  Computational Linguistics}, pages 7352--7364, Online, July. Association for
  Computational Linguistics.

\bibitem[\protect\citename{Balcerzak \bgroup et al.\egroup
  }2014]{balcerzak2014application}
Bartomiej Balcerzak, Wojciech Jaworski, and Adam Wierzbicki.
\newblock 2014.
\newblock Application of textrank algorithm for credibility assessment.
\newblock In {\em 2014 IEEE/WIC/ACM International Joint Conferences on Web
  Intelligence (WI) and Intelligent Agent Technologies (IAT)}, volume~1, pages
  451--454. IEEE.

\bibitem[\protect\citename{Barrios \bgroup et al.\egroup
  }2015]{barrios2015variations}
Federico Barrios, Federico L{\'o}pez, Luis Argerich, and Rosita Wachenchauzer.
\newblock 2015.
\newblock Variations of the similarity function of textrank for automated
  summarization.
\newblock In {\em Argentine Symposium on Artificial Intelligence (ASAI
  2015)-JAIIO 44 (Rosario, 2015)}.

\bibitem[\protect\citename{Bird and Loper}2004]{bird-loper-2004-nltk}
Steven Bird and Edward Loper.
\newblock 2004.
\newblock {NLTK}: The natural language toolkit.
\newblock In {\em Proceedings of the {ACL} Interactive Poster and Demonstration
  Sessions}, pages 214--217, Barcelona, Spain, July. Association for
  Computational Linguistics.

\bibitem[\protect\citename{Cao \bgroup et al.\egroup
  }2016]{cao-etal-2016-attsum}
Ziqiang Cao, Wenjie Li, Sujian Li, Furu Wei, and Yanran Li.
\newblock 2016.
\newblock {A}tt{S}um: Joint learning of focusing and summarization with neural
  attention.
\newblock In {\em Proceedings of {COLING} 2016, the 26th International
  Conference on Computational Linguistics: Technical Papers}, pages 547--556,
  Osaka, Japan, December. The COLING 2016 Organizing Committee.

\bibitem[\protect\citename{Dang}2005]{dang2005overview}
Hoa~Trang Dang.
\newblock 2005.
\newblock Overview of duc 2005.
\newblock In {\em Proceedings of the document understanding conference}, volume
  2005, pages 1--12.

\bibitem[\protect\citename{Daum{\'e}~III and
  Marcu}2006]{daume-iii-marcu-2006-bayesian}
Hal Daum{\'e}~III and Daniel Marcu.
\newblock 2006.
\newblock {B}ayesian query-focused summarization.
\newblock In {\em Proceedings of the 21st International Conference on
  Computational Linguistics and 44th Annual Meeting of the Association for
  Computational Linguistics}, pages 305--312, Sydney, Australia, July.
  Association for Computational Linguistics.

\bibitem[\protect\citename{Deguchi and
  Yamaguchi}2019]{deguchi-yamaguchi-2019-argument}
Mamoru Deguchi and Kazunori Yamaguchi.
\newblock 2019.
\newblock Argument component classification by relation identification by
  neural network and {T}ext{R}ank.
\newblock In {\em Proceedings of the 6th Workshop on Argument Mining}, pages
  83--91, Florence, Italy, August. Association for Computational Linguistics.

\bibitem[\protect\citename{Devlin \bgroup et al.\egroup
  }2019]{devlin-etal-2019-bert}
Jacob Devlin, Ming-Wei Chang, Kenton Lee, and Kristina Toutanova.
\newblock 2019.
\newblock {BERT}: Pre-training of deep bidirectional transformers for language
  understanding.
\newblock In {\em Proceedings of the 2019 Conference of the North {A}merican
  Chapter of the Association for Computational Linguistics: Human Language
  Technologies, Volume 1 (Long and Short Papers)}, pages 4171--4186,
  Minneapolis, Minnesota, June. Association for Computational Linguistics.

\bibitem[\protect\citename{Florescu and Caragea}2017]{florescu2017position}
Corina Florescu and Cornelia Caragea.
\newblock 2017.
\newblock A position-biased pagerank algorithm for keyphrase extraction.
\newblock In {\em Thirty-first AAAI conference on artificial intelligence}.

\bibitem[\protect\citename{Haveliwala}2003]{haveliwala2003topic}
Taher~H Haveliwala.
\newblock 2003.
\newblock Topic-sensitive pagerank: A context-sensitive ranking algorithm for
  web search.
\newblock {\em IEEE transactions on knowledge and data engineering},
  15(4):784--796.

\bibitem[\protect\citename{Hermann \bgroup et al.\egroup
  }2015]{hermann2015teaching}
Karl~Moritz Hermann, Tomas Kocisky, Edward Grefenstette, Lasse Espeholt, Will
  Kay, Mustafa Suleyman, and Phil Blunsom.
\newblock 2015.
\newblock Teaching machines to read and comprehend.
\newblock In {\em Advances in neural information processing systems}, pages
  1693--1701.

\bibitem[\protect\citename{Jianfei and Jiangzhen}2016]{jianfei2016using}
Ning Jianfei and Liu Jiangzhen.
\newblock 2016.
\newblock Using word2vec with textrank to extract keywords.
\newblock {\em Data Analysis and Knowledge Discovery}, 32(6):20--27.

\bibitem[\protect\citename{Kumar and Talukdar}2020]{kumar-talukdar-2020-nile}
Sawan Kumar and Partha Talukdar.
\newblock 2020.
\newblock {NILE} : Natural language inference with faithful natural language
  explanations.
\newblock In {\em Proceedings of the 58th Annual Meeting of the Association for
  Computational Linguistics}, pages 8730--8742, Online, July. Association for
  Computational Linguistics.

\bibitem[\protect\citename{Kushilevitz \bgroup et al.\egroup
  }2020]{kushilevitz-etal-2020-two}
Guy Kushilevitz, Shaul Markovitch, and Yoav Goldberg.
\newblock 2020.
\newblock A two-stage masked {LM} method for term set expansion.
\newblock In {\em Proceedings of the 58th Annual Meeting of the Association for
  Computational Linguistics}, pages 6829--6835, Online, July. Association for
  Computational Linguistics.

\bibitem[\protect\citename{Lin}2004]{lin-2004-rouge}
Chin-Yew Lin.
\newblock 2004.
\newblock {ROUGE}: A package for automatic evaluation of summaries.
\newblock In {\em Text Summarization Branches Out}, pages 74--81, Barcelona,
  Spain, July. Association for Computational Linguistics.

\bibitem[\protect\citename{Lu and Li}2020]{lu-li-2020-gcan}
Yi-Ju Lu and Cheng-Te Li.
\newblock 2020.
\newblock {GCAN}: Graph-aware co-attention networks for explainable fake news
  detection on social media.
\newblock In {\em Proceedings of the 58th Annual Meeting of the Association for
  Computational Linguistics}, pages 505--514, Online, July. Association for
  Computational Linguistics.

\bibitem[\protect\citename{Lundberg and Lee}2017]{lundberg2017unified}
Scott~M Lundberg and Su-In Lee.
\newblock 2017.
\newblock A unified approach to interpreting model predictions.
\newblock In {\em Advances in Neural Information Processing Systems}, pages
  4765--4774.

\bibitem[\protect\citename{Mallick \bgroup et al.\egroup
  }2019]{mallick2019graph}
Chirantana Mallick, Ajit~Kumar Das, Madhurima Dutta, Asit~Kumar Das, and Apurba
  Sarkar.
\newblock 2019.
\newblock Graph-based text summarization using modified textrank.
\newblock In {\em Soft Computing in Data Analytics}, pages 137--146. Springer.

\bibitem[\protect\citename{Metzler and Kanungo}2008]{metzler2008machine}
Donald Metzler and Tapas Kanungo.
\newblock 2008.
\newblock Machine learned sentence selection strategies for query-biased
  summarization.
\newblock In {\em Sigir learning to rank workshop}, pages 40--47. Citeseer.

\bibitem[\protect\citename{Mihalcea and
  Tarau}2004]{mihalcea-tarau-2004-textrank}
Rada Mihalcea and Paul Tarau.
\newblock 2004.
\newblock {T}ext{R}ank: Bringing order into text.
\newblock In {\em Proceedings of the 2004 Conference on Empirical Methods in
  Natural Language Processing}, pages 404--411, Barcelona, Spain, July.
  Association for Computational Linguistics.

\bibitem[\protect\citename{Mikolov \bgroup et al.\egroup
  }2013]{mikolov2013distributed}
Tomas Mikolov, Ilya Sutskever, Kai Chen, Greg~S Corrado, and Jeff Dean.
\newblock 2013.
\newblock Distributed representations of words and phrases and their
  compositionality.
\newblock In {\em Advances in neural information processing systems}, pages
  3111--3119.

\bibitem[\protect\citename{Page \bgroup et al.\egroup }1999]{ilprints422}
Lawrence Page, Sergey Brin, Rajeev Motwani, and Terry Winograd.
\newblock 1999.
\newblock The pagerank citation ranking: Bringing order to the web.
\newblock Technical Report 1999-66, Stanford InfoLab, November.
\newblock Previous number = SIDL-WP-1999-0120.

\bibitem[\protect\citename{Petasis and
  Karkaletsis}2016]{petasis2016identifying}
Georgios Petasis and Vangelis Karkaletsis.
\newblock 2016.
\newblock Identifying argument components through textrank.
\newblock In {\em Proceedings of the Third Workshop on Argument Mining
  (ArgMining2016)}, pages 94--102.

\bibitem[\protect\citename{Popat \bgroup et al.\egroup
  }2018]{popat-etal-2018-declare}
Kashyap Popat, Subhabrata Mukherjee, Andrew Yates, and Gerhard Weikum.
\newblock 2018.
\newblock {D}e{C}lar{E}: Debunking fake news and false claims using
  evidence-aware deep learning.
\newblock In {\em Proceedings of the 2018 Conference on Empirical Methods in
  Natural Language Processing}, pages 22--32, Brussels, Belgium,
  October-November. Association for Computational Linguistics.

\bibitem[\protect\citename{Poursabzi-Sangdeh \bgroup et al.\egroup
  }2018]{poursabzi2018manipulating}
Forough Poursabzi-Sangdeh, Daniel~G Goldstein, Jake~M Hofman, Jennifer~Wortman
  Vaughan, and Hanna Wallach.
\newblock 2018.
\newblock Manipulating and measuring model interpretability.
\newblock {\em arXiv preprint arXiv:1802.07810}.

\bibitem[\protect\citename{Rajpurkar \bgroup et al.\egroup
  }2018]{rajpurkar-etal-2018-know}
Pranav Rajpurkar, Robin Jia, and Percy Liang.
\newblock 2018.
\newblock Know what you don{'}t know: Unanswerable questions for {SQ}u{AD}.
\newblock In {\em Proceedings of the 56th Annual Meeting of the Association for
  Computational Linguistics (Volume 2: Short Papers)}, pages 784--789,
  Melbourne, Australia, July. Association for Computational Linguistics.

\bibitem[\protect\citename{Reddy \bgroup et al.\egroup
  }2019]{reddy-etal-2019-coqa}
Siva Reddy, Danqi Chen, and Christopher~D. Manning.
\newblock 2019.
\newblock {C}o{QA}: A conversational question answering challenge.
\newblock {\em Transactions of the Association for Computational Linguistics},
  7:249--266, March.

\bibitem[\protect\citename{Reimers and
  Gurevych}2019]{reimers-gurevych-2019-sentence}
Nils Reimers and Iryna Gurevych.
\newblock 2019.
\newblock Sentence-{BERT}: Sentence embeddings using {S}iamese {BERT}-networks.
\newblock In {\em Proceedings of the 2019 Conference on Empirical Methods in
  Natural Language Processing and the 9th International Joint Conference on
  Natural Language Processing (EMNLP-IJCNLP)}, pages 3982--3992, Hong Kong,
  China, November. Association for Computational Linguistics.

\bibitem[\protect\citename{Son and Shin}2018]{son2018music}
Jiyoung Son and Yongtae Shin.
\newblock 2018.
\newblock Music lyrics summarization method using textrank algorithm.
\newblock {\em Journal of Korea Multimedia Society}, 21(1):45--50.

\bibitem[\protect\citename{Strubell \bgroup et al.\egroup
  }2019]{strubell-etal-2019-energy}
Emma Strubell, Ananya Ganesh, and Andrew McCallum.
\newblock 2019.
\newblock Energy and policy considerations for deep learning in {NLP}.
\newblock In {\em Proceedings of the 57th Annual Meeting of the Association for
  Computational Linguistics}, pages 3645--3650, Florence, Italy, July.
  Association for Computational Linguistics.

\bibitem[\protect\citename{Thorne \bgroup et al.\egroup
  }2019]{thorne-etal-2019-generating}
James Thorne, Andreas Vlachos, Christos Christodoulopoulos, and Arpit Mittal.
\newblock 2019.
\newblock Generating token-level explanations for natural language inference.
\newblock In {\em Proceedings of the 2019 Conference of the North {A}merican
  Chapter of the Association for Computational Linguistics: Human Language
  Technologies, Volume 1 (Long and Short Papers)}, pages 963--969, Minneapolis,
  Minnesota, June. Association for Computational Linguistics.

\bibitem[\protect\citename{Wan}2008]{wan2008using}
Xiaojun Wan.
\newblock 2008.
\newblock Using only cross-document relationships for both generic and
  topic-focused multi-document summarizations.
\newblock {\em Information Retrieval}, 11(1):25--49.

\bibitem[\protect\citename{Wang \bgroup et al.\egroup }2007]{wang2007learning}
Changhu Wang, Feng Jing, Lei Zhang, and Hong-Jiang Zhang.
\newblock 2007.
\newblock Learning query-biased web page summarization.
\newblock In {\em Proceedings of the sixteenth ACM conference on Conference on
  information and knowledge management}, pages 555--562.

\bibitem[\protect\citename{Wen \bgroup et al.\egroup }2016]{wen2016research}
Yujun Wen, Hui Yuan, and Pengzhou Zhang.
\newblock 2016.
\newblock Research on keyword extraction based on word2vec weighted textrank.
\newblock In {\em 2016 2nd IEEE International Conference on Computer and
  Communications (ICCC)}, pages 2109--2113. IEEE.

\bibitem[\protect\citename{Zhao \bgroup et al.\egroup }2009]{zhao2009using}
Lin Zhao, Lide Wu, and Xuanjing Huang.
\newblock 2009.
\newblock Using query expansion in graph-based approach for query-focused
  multi-document summarization.
\newblock {\em Information processing \& management}, 45(1):35--41.

\end{thebibliography}

\end{document}